\documentclass[11pt]{article}
\usepackage[utf8]{inputenc}
\usepackage[T1]{fontenc}
\usepackage{amsmath,amssymb}
\usepackage{graphicx}
\usepackage{booktabs}
\usepackage{hyperref}
\usepackage{natbib}
\usepackage[margin=1in]{geometry}
\usepackage{xcolor}
\usepackage{enumitem}
\usepackage{float}

\hypersetup{
    colorlinks=true,
    linkcolor=blue,
    citecolor=blue,
    urlcolor=blue
}

\title{Prompt Architecture Determines Reasoning Quality:\\A Variable Isolation Study on the Car Wash Problem}

\author{Heejin Jo\\
  Independent Researcher\\
  \texttt{info@birth2death.com}\\
  \url{https://www.heejinjo.me}
}

\date{February 25, 2026}

\begin{document}
\maketitle

\begin{abstract}
The car wash problem asks a simple question: ``I want to wash my car. The car wash is 100 meters away. Should I walk or drive?'' Every major LLM tested---Claude, GPT-4, Gemini---recommended walking. The correct answer is to drive, because the car itself must be at the car wash.

We ran a variable isolation study to determine which prompt architectural layer resolves this failure. Six conditions were tested, 20 trials each, on Claude Sonnet 4.5. A bare prompt with no system instructions scored 0\%. Adding a role definition alone also scored 0\%. A STAR reasoning framework (Situation, Task, Action, Result) reached 85\%. User profile injection with physical context---car model, location, parking status---reached only 30\%. STAR combined with profile injection reached 95\%. The full stack combining all layers scored 100\%.

The central finding is that structured reasoning outperformed direct context injection by a factor of 2.83$\times$ (Fisher's exact test, $p = 0.001$). STAR forces the model to articulate the task goal before generating a conclusion, which surfaces the implicit physical constraint that context injection leaves buried. The addition of a sixth condition resolved a confound in the original five-condition design by isolating per-layer contributions: STAR accounts for +85pp, profile adds +10pp, and RAG provides the final +5pp to reach perfect reliability.
\end{abstract}

\section{Introduction}

The question originated on Mastodon, posted by Kevin (@knowmadd)\footnote{\url{https://mastodon.world/@knowmadd/116072773118828295}}: ``I want to wash my car. The car wash is 50 meters away. Should I walk or drive?'' He tested Perplexity, ChatGPT, Claude, and Mistral. All four said walk. The correct answer is drive---you cannot wash a car that is not there.

The post reached Hacker News and accumulated 1,499 points and 943 comments.\footnote{\url{https://news.ycombinator.com/item?id=47031580}} Discussion centered on three themes: that LLMs cannot infer implicit prerequisites humans take for granted; the classical frame problem \citep{mccarthy1969}, where models fail to identify which unstated facts matter; and whether the gap between natural and structured communication is something models can be made to bridge. Ryan Allen later published a benchmark repository\footnote{\url{https://github.com/ryan-allen/car-wash-evals}} to measure this failure across models.

We encountered this problem through InterviewMate, a real-time interview coaching system. During a routine test session, the system answered ``drive'' while every standalone LLM we tested said ``walk.'' We did not expect this. InterviewMate's system prompt has multiple layers---role definition, a STAR reasoning framework, user profile data, and RAG context retrieval---and we had no way to tell which layer produced the correct answer. The result was interesting, but we could not explain it, and a result you cannot explain is not one you can build on.

So we designed a variable isolation experiment. Instead of asking why LLMs fail at this problem---a question the Hacker News thread had already covered extensively---we asked which specific prompt layer fixes it within a single model.

This question has practical weight. InterviewMate operates during live interviews. The system must interpret the interviewer's question, retrieve relevant context from the user's stored profile and STAR stories, and deliver a structured answer in real time. The prompt architecture follows this sequence:

\begin{center}
Role definition $\rightarrow$ STAR reasoning framework $\rightarrow$ User profile (vector DB) $\rightarrow$ RAG context retrieval
\end{center}

We needed to know where the reasoning quality actually comes from in this stack. The car wash problem gave us a clean instrument for testing it: one correct answer, implicit constraint reasoning required, and simple enough to isolate variables without confounds.

\section{Related Work}

\textbf{The Car Wash Benchmark.} Ryan Allen published a formal evaluation repository (ryan-allen/car-wash-evals) that measures car wash problem failure rates across models. His work established a baseline: most frontier LLMs fail on first pass. Our study starts from this baseline but asks a different question---not which models fail, but which prompt layers fix the failure within a single model.

\textbf{The Frame Problem.} \citet{mccarthy1969} described the frame problem in classical AI: a system must determine which facts remain unchanged when an action occurs, but has no principled method for knowing which unstated facts are relevant. The car wash problem is a clean modern example. The car's location is never mentioned in the question. A human infers immediately that the car is at home. The model does not.

\textbf{Chain-of-Thought Prompting.} \citet{wei2022} showed that prompting models to reason step-by-step significantly improves performance on multi-step tasks. Our STAR result---85\% versus a 0\% baseline---is broadly consistent with this. Structured reasoning forces the model to work through intermediate steps rather than jumping to a conclusion.

\textbf{Self-Consistency.} \citet{wang2023} extended chain-of-thought by sampling multiple reasoning paths and selecting the most frequent answer. Our experimental design differs---we measure pass rates across 20 independent trials per condition rather than sampling multiple paths within a single trial---but the underlying observation is related. Reasoning quality varies across runs even with identical prompts, and aggregation can reveal the underlying success rate.

\textbf{ReAct.} \citet{yao2023react} interleaved reasoning steps with action steps, allowing models to gather external information during the reasoning process. STAR is structurally simpler: it does not involve external actions. But both frameworks share the principle that forcing explicit intermediate steps before a conclusion changes the output distribution.

\textbf{Tree of Thoughts.} \citet{yao2023tree} proposed exploring multiple reasoning branches and evaluating them before committing to an answer. We did not implement branching in our study, but the Tree of Thoughts result supports a broader point: the structure imposed on a model's reasoning process has measurable effects on accuracy, independent of the information available.

\textbf{Where STAR Fits.} STAR (Situation, Task, Action, Result) is not a research contribution of this paper. It is a standard interview preparation framework that we repurposed as a prompt structure. What makes it interesting in this context is the Task step specifically. Chain-of-thought asks the model to reason step by step. STAR forces the model to name what it is trying to accomplish before it begins reasoning about how. The distinction is between general sequential reasoning and explicit goal articulation. Section~\ref{sec:why_star} examines this mechanism in detail.

\section{Methodology}

This is a pilot study. The sample size ($n=20$ per condition, 120 total API calls) is sufficient to identify behavioral patterns and directional differences between conditions. We report pass rates, observed patterns, and statistical significance testing where applicable.

\subsection{Experimental Setup}

All trials used \texttt{claude-sonnet-4-5-20250929} via the Anthropic Python SDK. This version was chosen for consistency with the ryan-allen/car-wash-evals baseline. Hyperparameters were fixed: \texttt{max\_tokens} at 512, temperature at 0.7. We chose 0.7 rather than 0 because deterministic decoding would collapse results toward binary outcomes, making per-layer contributions impossible to distinguish. The variance introduced by temperature 0.7 across 20 runs lets us measure probabilistic pass rates.

The prompt was modified from the original Mastodon question (50 meters) to 100 meters: ``I want to wash my car. The car wash is 100 meters away. Should I walk or drive?'' This follows the ryan-allen/car-wash-evals benchmark. At 100 meters, the distance is more ambiguous than 50 meters, which may make ``drive'' slightly more intuitive and the task slightly easier than the original (see Limitations). Despite this, all baseline conditions without structured reasoning still defaulted to ``walk,'' suggesting that distance alone does not trigger correct implicit reasoning.

For any trial that failed or produced an ambiguous result, a challenge prompt was given: ``How will I get my car washed if I am walking?'' This measures self-correction capability, which we report as Recovery Rate.

\subsection{Variable Isolation Design}

Six conditions were tested. Each ran 20 independent trials.

\begin{itemize}[leftmargin=2em]
\item \textbf{Condition A (Bare):} No system prompt. Pure baseline to measure the model's default behavior.
\item \textbf{Condition B (Role Only):} An expert advisor persona was injected as the system prompt.
\item \textbf{Condition C (Role + STAR):} The STAR reasoning framework was added on top of the role, requiring the model to articulate Situation, Task, Action, and Result in sequence.
\item \textbf{Condition D (Role + Profile):} Instead of STAR, physical user context was injected---name, location, vehicle model, current situation. Conditions C and D branch from B in parallel. They are not cumulative.
\item \textbf{Condition E (Full Stack):} All layers combined---Role, STAR, Profile, and simulated RAG context.
\item \textbf{Condition F (Role + STAR + Profile):} STAR and profile combined, without RAG context. Added to resolve the E\_full\_stack confound identified during initial analysis, isolating per-layer contributions between profile and RAG.
\end{itemize}

The design isolates STAR (C vs B), profile injection (D vs B), profile's marginal contribution on top of STAR (F vs C), RAG's marginal contribution (E vs F), and the full interaction (E vs C, D, and F).

\subsection{Scoring Methodology}

The first evaluation run used bare word matching, checking for ``walk'' or ``drive'' in the response. This failed. Every response that discussed both options was classified as ambiguous, which produced 0\% pass rates across all conditions regardless of actual recommendation. The scorer was not measuring intent.

Run 2 replaced word matching with intent-based pattern matching: 14 pass patterns detecting drive recommendations (e.g., \texttt{\textbackslash bshould\textbackslash s+drive\textbackslash b}) and 9 fail patterns detecting walk recommendations (e.g., \texttt{\textbackslash brecommend\textbackslash s+walking\textbackslash b}).

One implementation detail: markdown bold markup had to be stripped before matching. Claude frequently writes ``should \textbf{walk}'' and the asterisks break whitespace-based regex patterns. When both pass and fail patterns matched in a single response, a dominance ratio determined the result. A 2:1 threshold was required for a definitive score; otherwise the trial was flagged ambiguous. Statistical significance testing was performed using \texttt{scipy.stats.fisher\_exact} (SciPy 1.12, two-tailed).

\section{Results}

\subsection{Primary Pass Rates}

\begin{table}[H]
\centering
\begin{tabular}{llccc}
\toprule
\textbf{Condition} & \textbf{Components} & \textbf{Pass Rate} & \textbf{Recovery} & \textbf{Med. Latency} \\
\midrule
A\_bare & No system prompt & 0\% (0/20) & 95\% & 4,649ms \\
B\_role\_only & Role only & 0\% (0/20) & 100\% & 7,550ms \\
C\_role\_star & Role + STAR & 85\% (17/20) & 67\% & 7,851ms \\
D\_role\_profile & Role + Profile & 30\% (6/20) & 100\% & 8,837ms \\
F\_role\_star\_profile & Role + STAR + Profile & 95\% (19/20) & 0\% (0/1) & 9,056ms \\
E\_full\_stack & All combined & 100\% (20/20) & n/a & 8,347ms \\
\bottomrule
\end{tabular}
\caption{Pass rates across six experimental conditions. Pass means the first response recommends driving. Recovery means the model self-corrects after the challenge prompt.}
\label{tab:results}
\end{table}

\subsection{Key Findings}

\textbf{Finding 1: Reasoning structure outperforms context injection by 2.83$\times$.} C\_role\_star reached 85\%. D\_role\_profile reached 30\%. Both branch from the same baseline (B, 0\%). The STAR framework forces the model to name the task before generating a conclusion. Profile injection gives the model physical facts---car model, location, parking status---but does not force it to process those facts in any particular order. The model can receive all the right information and still take a shortcut past it. The difference between C (85\%) and D (30\%) was statistically significant (Fisher's exact test, two-tailed, $p = 0.001$, odds ratio $= 13.22$), confirming that the observed 2.83$\times$ advantage of structured reasoning over context injection is unlikely to be attributable to chance even at this sample size.

\textbf{Finding 2: Per-layer contributions are now isolated.} The addition of condition F resolves the confound in the original five-condition design. The progression from C to E can now be decomposed:

\begin{itemize}[leftmargin=2em]
\item STAR alone: +85pp (0\% to 85\%)
\item Profile on top of STAR: +10pp (85\% to 95\%)
\item RAG on top of STAR + Profile: +5pp (95\% to 100\%)
\end{itemize}

Profile's marginal contribution (+10pp) is twice that of RAG (+5pp). Both are necessary for perfect reliability, but STAR accounts for the overwhelming majority of the improvement.

\textbf{Finding 3: The recovery paradox.} C\_role\_star had the highest first-pass accuracy but the lowest recovery rate (67\%). Conditions A, B, and D all recovered at 95--100\%. F\_role\_star\_profile showed an even more extreme version of this pattern: its single failure (1/20) did not recover at all (0\% recovery rate). Section~\ref{sec:recovery} discusses the mechanism.

\textbf{Finding 4: Role definition alone does nothing.} B\_role\_only scored 0\%, identical to A\_bare. The expert advisor persona made the model write longer responses (median latency 7,550ms vs 4,649ms) but did not change the answer.

\subsection{Failure Mode Taxonomy}

Three failure patterns appeared across all failed primary trials:

\textbf{Type 1---Distance Heuristic} ($\sim$70\% of failures). The model treats the question as a distance optimization problem. ``100 meters is a 1--2 minute walk.'' It never considers what needs to be at the destination.

\textbf{Type 2---Environmental Rationalization} ($\sim$20\%). The model builds secondary justifications around the wrong answer. ``Walking saves fuel and is better for the environment.''

\textbf{Type 3---Ironic Self-Awareness} ($\sim$10\%). The model acknowledges that the car needs to be at the car wash, then still recommends walking. One B\_role\_only response said: ``You can drive your car through the wash bay when you arrive''---as if the car would get there on its own.

\subsection{Latency Analysis}

\begin{table}[H]
\centering
\begin{tabular}{lcc}
\toprule
\textbf{Condition} & \textbf{Median Latency} & \textbf{vs Baseline} \\
\midrule
A\_bare & 4,649ms & --- \\
B\_role\_only & 7,550ms & +62\% \\
C\_role\_star & 7,851ms & +69\% \\
D\_role\_profile & 8,837ms & +90\% \\
F\_role\_star\_profile & 9,056ms & +95\% \\
E\_full\_stack & 8,347ms & +80\% \\
\bottomrule
\end{tabular}
\caption{Median response latencies. E\_full\_stack is faster than D and F despite having more context, suggesting that model confidence reduces deliberation time.}
\label{tab:latency}
\end{table}

\section{Discussion}

\subsection{Why STAR Works: The Task Step}
\label{sec:why_star}

The mechanism is in the Task step. Without STAR, the model goes straight from ``100 meters'' to the distance heuristic to ``walk.'' The purpose of the trip---washing the car---is in the input, but the model has no obligation to process it before reaching a conclusion.

STAR changes the generation sequence. The model must fill in:

\begin{quote}
Situation: I want to wash my car. The car wash is 100 meters away.\\
Task: \_\_\_
\end{quote}

This creates a fork. If the model writes ``Task: Get your car to the car wash,'' the car becomes the subject of the goal statement. Drive follows naturally. If the model writes ``Task: Get yourself and your car to the car wash efficiently,'' the person re-enters as the subject, and walk becomes plausible again.

The per-trial data confirms this. All 17 passing trials in C\_role\_star had Task statements where the car was the primary subject. All 3 failures framed the task around the person.

Once the model generates ``Task: Get your car to the car wash,'' every token that follows is conditioned on that text. The implicit constraint---the car must physically be there---is now explicit in the context window. STAR does not give the model new information. It makes the model write down what it already has before moving on.

\subsection{Why Profile Injection Alone Falls Short}

D\_role\_profile gave the model everything it needed: Sarah drives a 2022 Honda Civic, it is parked in the driveway, she is at home. This is enough to answer correctly. The pass rate was 30\%.

The problem is not about missing information. The model has the facts. But having facts in the context window does not mean the model will use them at the right moment. Without a reasoning structure, the model still takes the shortest path from input to output. ``100 meters'' triggers the distance heuristic and the conclusion lands before the car's location ever gets pulled into the reasoning chain.

\subsection{The Recovery Paradox}
\label{sec:recovery}

C\_role\_star scored 85\% on first pass but only 67\% on recovery. A\_bare and B\_role\_only scored 0\% on first pass but recovered at 95--100\%. F\_role\_star\_profile showed an even more extreme version: 95\% first-pass accuracy but 0\% recovery on its single failure.

When conditions A and B fail, they fail with loose, unstructured responses. A challenge prompt can redirect easily because there is nothing anchoring the wrong answer. When C or F fails, it fails with a full STAR-structured argument. The model has already walked through Situation, Task, Action, and Result, and produced a coherent case for walking. Correcting course means contradicting a structured argument the model just made.

The mechanism is token-level, not psychological. Prior generated text constrains subsequent generation. This has a practical consequence: if an initial response used structured reasoning, the follow-up correction needs to target the specific step that went wrong---in this case, the Task formulation.

\subsection{The Profile-RAG Decomposition}

The addition of condition F resolves a confound present in the original five-condition design. Previously, the jump from 85\% (C) to 100\% (E) could not be attributed to any single layer because E added both profile and RAG simultaneously.

With F at 95\%, the decomposition is now clear:

\begin{table}[H]
\centering
\begin{tabular}{lcc}
\toprule
\textbf{Layer Added} & \textbf{Increment} & \textbf{Cumulative} \\
\midrule
STAR (C vs B) & +85pp & 85\% \\
Profile (F vs C) & +10pp & 95\% \\
RAG (E vs F) & +5pp & 100\% \\
\bottomrule
\end{tabular}
\caption{Per-layer marginal contributions to pass rate.}
\label{tab:decomposition}
\end{table}

Profile contributes twice as much as RAG in the final stretch. The mechanism is likely that profile grounds the STAR framework in concrete physical details (a specific car, a specific location), which reduces the probability of the Task step being formulated abstractly. RAG adds situational context (the car needs washing after a road trip) that eliminates the remaining edge case.

\subsection{Open Questions}

This study measures behavior at the prompt layer. We can see that STAR produces 85\% and profile injection produces 30\%. We do not know what happens inside the model to produce this difference.

Which attention heads activate differently when the Task step is present? Does forcing goal articulation change the activation pattern in a way that is consistent across prompts, or is it specific to this question? Would the same STAR structure produce the same lift on GPT-4 or Gemini, or is the effect tied to Claude's training?

These are mechanistic interpretability questions. What we have is a behavioral result that any interpretability study could use as a starting point: the same model, the same question, two prompt conditions, a 55 percentage point gap.

\section{Limitations}

\textbf{Single model.} Every trial used \texttt{claude-sonnet-4-5-20250929}. Whether these patterns hold across GPT-4o, Gemini, or Mistral is unknown.

\textbf{Single task.} One question, one correct answer. The car wash problem tests implicit physical constraint reasoning specifically. Whether STAR produces similar gains on temporal constraints, social context inference, or causal chain reasoning has not been tested.

\textbf{Sample size.} 20 runs per condition. The Fisher's exact test confirms that the C vs D difference is statistically significant ($p = 0.001$), but confidence intervals around individual pass rates remain wide.

\textbf{Temperature.} 0.7 was chosen as a reasonable default for introducing variance. We did not sweep across temperature values.

\textbf{Distance modification.} The original Mastodon question used 50 meters. We used 100, following the benchmark. The longer distance might make the task slightly easier than the original.

\textbf{DeepSeek version.} The model used for pre-experiment prediction was DeepSeek, but we did not record the exact version or endpoint.

\textbf{Latency overhead.} STAR-structured prompts increased median response time by about 69\% over baseline (7,851ms vs 4,649ms).

\textbf{Challenge prompt bias.} The challenge prompt (``How will I get my car washed if I am walking?'') is leading. A neutral challenge like ``Are you sure?'' would better isolate self-correction ability.

\textbf{F condition timing.} Condition F was added after the initial five-condition experiment, approximately six days later. While the same model version, hyperparameters, and scoring methodology were used, we cannot rule out that API-level changes may have introduced minor behavioral differences.

\section{Conclusion}

We started this study because we had a result we could not explain. Our system answered a question correctly when other LLMs did not, and we did not know which part of the system was responsible.

The experiment gave us a clear answer:

\begin{table}[H]
\centering
\begin{tabular}{lcc}
\toprule
\textbf{Layer} & \textbf{Pass Rate} & \textbf{Marginal Contribution} \\
\midrule
Baseline (A, B) & 0\% & --- \\
STAR (C) & 85\% & +85pp \\
Profile alone (D) & 30\% & +30pp (without STAR) \\
STAR + Profile (F) & 95\% & +10pp (on top of STAR) \\
STAR + Profile + RAG (E) & 100\% & +5pp (on top of STAR + Profile) \\
\bottomrule
\end{tabular}
\caption{Complete layer progression from baseline to perfect reliability.}
\label{tab:conclusion}
\end{table}

STAR reasoning accounts for the overwhelming majority of the improvement. Profile and RAG are necessary for perfect reliability but insufficient on their own. The mechanism appears to be goal articulation: when the model is forced to write down what it is trying to accomplish before it starts reasoning about how, implicit constraints surface as explicit text. Once they are explicit, autoregressive generation conditions on them.

There is a broader point here. A common pattern in applied AI is to solve reasoning failures by adding more context---more facts, more profile data, more retrieved documents. Our results suggest this is the wrong first move. How the model processes information matters more than how much information it receives. Profile injection with all the right facts scored 30\%. Structured reasoning with no additional facts scored 85\%. The difference was statistically significant ($p = 0.001$).

Or to put it less formally: intelligence is not about how much you hold in your head. It is about knowing to pick up the keys before you leave the house.

\bibliographystyle{plainnat}

\end{document}